\documentclass[conference]{IEEEtran}
\IEEEoverridecommandlockouts

\usepackage{cite}
\usepackage{amsmath,amssymb,amsfonts}
\usepackage{algorithmic}
\usepackage[ruled,vlined]{algorithm2e}
\usepackage{graphicx}
\usepackage{textcomp}
\usepackage{xcolor}
\usepackage{booktabs}
\usepackage{multirow}
\usepackage{makecell}
\usepackage{hhline}
\usepackage{svg}
\usepackage{subcaption}

\def\BibTeX{{\rm B\kern-.05em{\sc i\kern-.025em b}\kern-.08em
    T\kern-.1667em\lower.7ex\hbox{E}\kern-.125emX}}
    
\begin{document}

\title{S4oP: Operator-level Pruning of Structured State Space Models for Resource-Constrained Devices}
    
\author{
\IEEEauthorblockN{Marco Deano}
\IEEEauthorblockA{\textit{University of Verona} \\
Italy \\
marco.deano@univr.it}
\and
\IEEEauthorblockN{Filippo Ziche}
\IEEEauthorblockA{\textit{University of Verona} \\
Italy \\
filippo.ziche@univr.it}
\and
\IEEEauthorblockN{Nicola Bombieri}
\IEEEauthorblockA{\textit{University of Verona} \\
Italy \\
nicola.bombieri@univr.it}
}

\maketitle

\begin{abstract}
Structured State Space Models (SSMs), including the S4 and S4D architectures, have recently emerged as powerful alternatives to attention-based models for capturing long-range dependencies in sequential data. Despite their strong empirical performance, deploying these models in time- and resource-constrained settings remains challenging due to their computational and memory demands. In this paper, we propose a novel incremental, operator-level pruning approach for S4- and S4D-based models that significantly reduces inference cost while preserving predictive performance. 

To the best of our knowledge, this is the first work to systematically investigate deployment-oriented operator-level pruning of parallel SSM channels in S4 and S4D architectures, explicitly targeting the trade-off between predictive performance and real inference latency on resource-constrained devices.
Our method progressively prunes model operators by interleaving structured masking with fine-tuning, while jointly monitoring accuracy and inference latency. We implement this approach within a unified training and evaluation framework that enables systematic exploration of efficiency-accuracy trade-offs.


Experiments across multiple datasets show that pruning up to 30\% of the model operators generally preserves the predictive performance of the original models, with only marginal accuracy degradation. In some cases, the pruning ratio can be increased up to 70\% while still maintaining comparable performance in predictive quality,  while substantially reducing inference latency.
These results demonstrate that structured operator pruning is an effective and previously unexplored strategy for improving the efficiency of SSMs and facilitate their deployment in practical, resource-constrained scenarios.

\end{abstract}


\section{Introduction}\label{SEC:INTRO}
\begin{figure}[h!]
    \centering
    \includegraphics[width=0.95\linewidth]{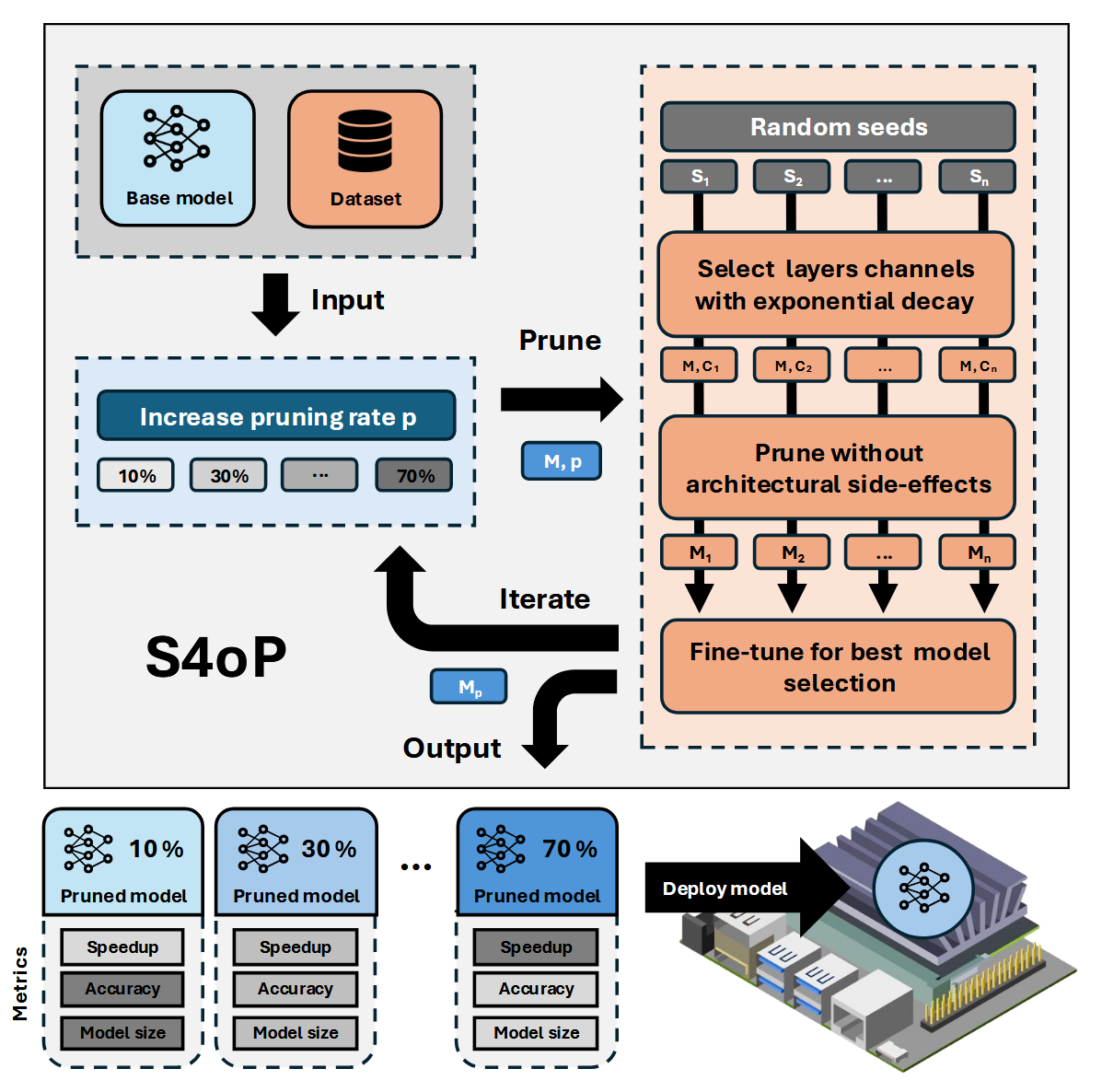}
    \caption{\textit{Overview of the proposed pruning approach.}}
    \label{fig:flow_chart}
    \vspace{-0.5cm}
\end{figure}

Structured State Space Models (SSMs), and in particular S4~\cite{s4} and S4D~\cite{s4d}, have recently emerged as effective architectures for sequential signal processing and time-series analysis, offering competitive accuracy and efficient modeling of long-range dependencies. Their complexity makes them appealing for deployment in resource-constrained systems, where latency, memory, and energy efficiency are critical.

However, despite their efficiency, deploying S4-based models on edge and IoT devices remains challenging due to limited hardware resources and inference constraints \cite{CHEN2026108459, Fu202310373}. This motivates the need for compression strategies able to reduce model cost while preserving predictive performance.

Model pruning is a widely adopted compression technique for reducing neural network complexity by removing redundant components \cite{survey_pruning}. While extensively studied for convolutional and transformer-based models, pruning remains largely unexplored for SSMs, particularly in terms of its effect on both accuracy and inference efficiency.

To address this gap, we propose $S4oP$, a pruning framework for S4 and S4D architectures based on incremental operator-level pruning (Figure~\ref{fig:flow_chart}). Instead of removing individual parameters, our method selectively disables computational operators while preserving tensor semantics by directly forwarding their inputs to outputs. Pruning is interleaved with fine-tuning and continuous evaluation.

We evaluate the proposed framework on multiple datasets and tasks, analyzing the impact of pruning on accuracy, latency, and memory usage. Results show that pruning around 30\% of model channels generally preserves baseline accuracy while significantly reducing inference latency. In several cases, more aggressive pruning (up to 50-70\%) maintains or even improves predictive performance with similar latency gains.

These results show that deployment-aware pruning is a promising direction for enabling efficient SSM inference in resource-constrained environments.

\section{Background and related works}\label{SEC:RELATED}
\subsection{Structured State Space Models}

Structured State Space Models (SSMs) model sequential data through latent linear dynamics, making them effective for capturing long-range dependencies in sequence tasks \cite{Gu2021572}. In discrete form, an SSM updates a latent state \(x_t\) from the input \(u_t\) and produces an output \(y_t\):
\[
x_{t+1} = \mathbf{A}x_t + \mathbf{B}u_t, \qquad
y_t = \mathbf{C}x_t + \mathbf{D}u_t.
\]
Unlike standard recurrent models, SSMs admit an equivalent convolutional formulation, enabling parallel computation over long sequences. In this way, the output sequence can be expressed as a convolution between the input and a kernel determined by the system parameters as
\[
y = \mathbf{\bar{K}} \ast u,
\]
enabling parallel computation over long sequences and overcoming the sequential bottleneck of recurrent models.

S4~\cite{s4} makes SSMs practical for deep learning by introducing a structured parameterization of the state matrix that enables efficient convolution kernel computation. S4D~\cite{s4d} further simplifies this formulation through a diagonal approximation, reducing computational cost while retaining competitive accuracy.

\begin{figure}[t]
    \centering
    \includegraphics[width=0.95\linewidth]{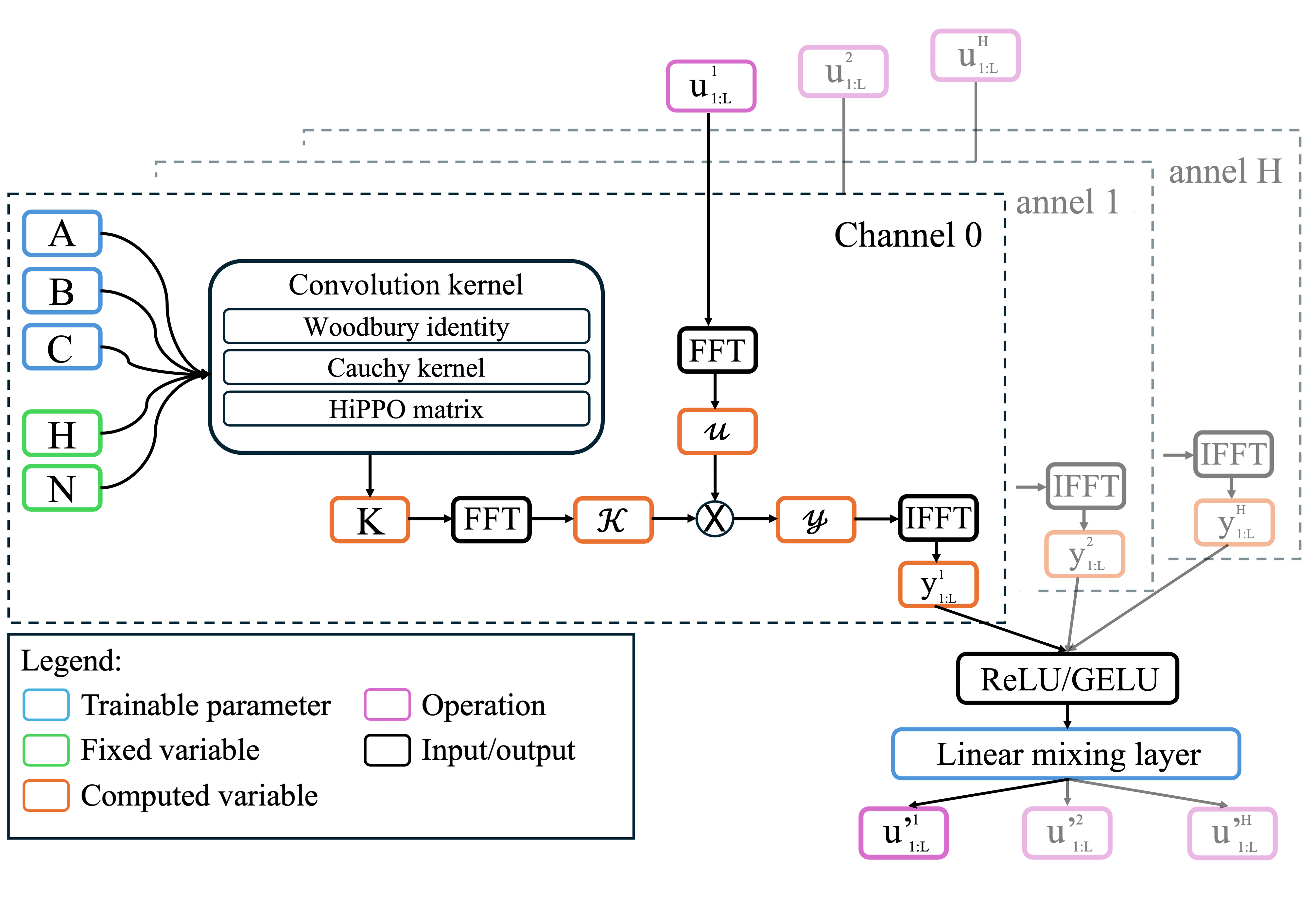}
    \caption{\textit{Architecture of an S4 layer.}}
    \label{fig:s4_layer}
    \vspace{-10pt}
\end{figure}

In deep architectures, S4 and S4D layers apply multiple independent state-space operators in parallel across feature channels, followed by learned mixing layers (Figure~\ref{fig:s4_layer}).

\subsection{Model pruning and deployment-aware optimization}

Model pruning reduces network complexity by removing redundant components while preserving predictive performance, making it a key technique for efficient deployment \cite{survey_pruning,deep_compression}. Structured pruning is particularly effective in practice, as removing groups of parameters yields more reliable acceleration on hardware than unstructured sparsity \cite{filter_pruning,channel_pruning}. Iterative pruning with interleaved fine-tuning further improves robustness by allowing the model to adapt after each pruning step \cite{iterative_pruning,rewinding_vs_finetuning}.

Recent work has shown that fine-grained structured pruning can achieve better efficiency-accuracy trade-offs than coarse layer removal, as demonstrated for Transformers by BlockPruner \cite{block_pruner} and for selective SSMs by Mamba-Shedder \cite{mamba_shedder}. 
Closer to our setting, \textit{Gwak et al.} \cite{LAST} propose a layer-adaptive method that reduces the state dimension of S4/S5-style SSMs, targeting redundancy in the internal state representation. In contrast, our work targets redundancy at the level of parallel SSM operators in S4 and S4D architectures, enabling deployment-oriented channel removal and reductions in inference latency without modifying the underlying state-space formulation.

In S4 and S4D, each channel corresponds to an independent state-space operator executed in parallel. This property naturally enables operator-level pruning, where selected channels are removed and bypassed by directly forwarding their input. Building on this observation, our method targets channel-level redundancy in S4 and S4D, enabling finer control of the efficiency-accuracy trade-off.

\section{The S4oP approach}\label{SEC:METHOD}
S4 and S4D layers are well suited to structured pruning because each channel implements an independent SSM operator, executed in parallel and later combined through linear mixing and residual connections (Figure~\ref{fig:s4_layer}). This structure enables operator-level pruning by removing entire SSM channels while preserving tensor dimensions: pruned channels are replaced with identity mappings, so downstream layers remain unchanged. Unlike parameter-level sparsification, this avoids disrupting kernel computation and FFT-based convolutions.

\subsection{Channel-level pruning}

S4oP adopts an incremental greedy strategy (Figure~\ref{fig:flow_chart}) that explores progressively higher pruning rates. Starting from the baseline model, channels are removed in stages and the best model at each step becomes the starting point for the next, ensuring nested and stable pruning configurations.

For each pruning rate $p$, multiple random pruning seeds are evaluated. Each seed samples a different set of channels, applies pruning, fine-tunes the resulting model for $e$ epochs, and evaluates it on the target task. The best-performing candidate is retained. This selection reduces sensitivity to unfavorable random configurations while keeping the search cost tractable.

Channel allocation across layers follows the depth-aware policy of Algorithm~\ref{algo:depth-aware_channel_selection_algorithm}. Given a target pruning rate, channels are distributed across layers with exponentially increasing weights, pruning deeper layers more aggressively while preserving earlier ones. At least one SSM operator is retained per layer to preserve functional capacity.

\begin{algorithm}[t]
    \SetAlgoLined
    \LinesNumbered
    \KwData{number of layers $n$, pruning rates $\mathcal{P}$, channels per layer $\mathit{H}$}
    \KwResult{channels to prune per layer for each rate}
    \ForEach{$p \in \mathcal{P}$}{
        $\mathcal{L} \gets \{1,\dots,n\}$; $\mathcal{C}_{p} \gets \{0 \mid e \in \mathcal{L}\}$\;
        $C_{target} \gets p \cdot H \cdot n$; $C_{rem} \gets C_{target}$\;
        \While{$C_{rem} > 0$}{
            $\mathcal{W} \gets \{2^i \mid i \in \mathcal{L}\}$; normalize($\mathcal{W}$)\;
            \ForEach{$i \in \mathcal{L}$}{
                $C_{alloc} \gets \mathcal{W}[i] \cdot C_{rem}$\;
                \eIf{$\mathcal{C}_{p}[i] + C_{alloc} \ge H-1$}{
                    $\mathcal{C}_{p}[i] \gets H-1$; $\mathcal{L} \gets \mathcal{L}\setminus\{i\}$\;
                }{
                    $\mathcal{C}_{p}[i] \gets \mathcal{C}_{p}[i] + C_{alloc}$\;
                }
            }
            $C_{rem} \gets C_{target} - sum(\mathcal{C}_p)$\;
        }
        yield $\mathcal{C}_{p}$\;
    }
    \caption{Depth-aware channel allocation.}
    \label{algo:depth-aware_channel_selection_algorithm}
\end{algorithm}

\subsection{Rationale for random channel selection}

Rather than ranking channels by importance, we adopt random channel selection. In S4 and S4D, the key factor governing performance is how many channels are pruned per layer, not which specific channels are removed, 
while fine-tuning effectively recovers from local perturbations. Prior work reports similar behavior in other architectures, where the pruned structure often matters more than the specific inherited weights \cite{rethinking_pruning,random_pruning}.

This avoids costly channel-scoring procedures and the combinatorial complexity of selecting optimal subsets, while retaining strong empirical performance and substantially reducing pruning cost.

\section{Experimental results}\label{SEC:EXPER}
We evaluate $S4oP$ on four benchmarks and report predictive performance (accuracy or F1-score) against pruning rate, together with inference latency and parameter reduction.

\begin{figure}[t!]
    \vspace{-10pt}
    \begin{subfigure}{0.49\textwidth}
        \centering
        \includegraphics[width=0.95\linewidth]{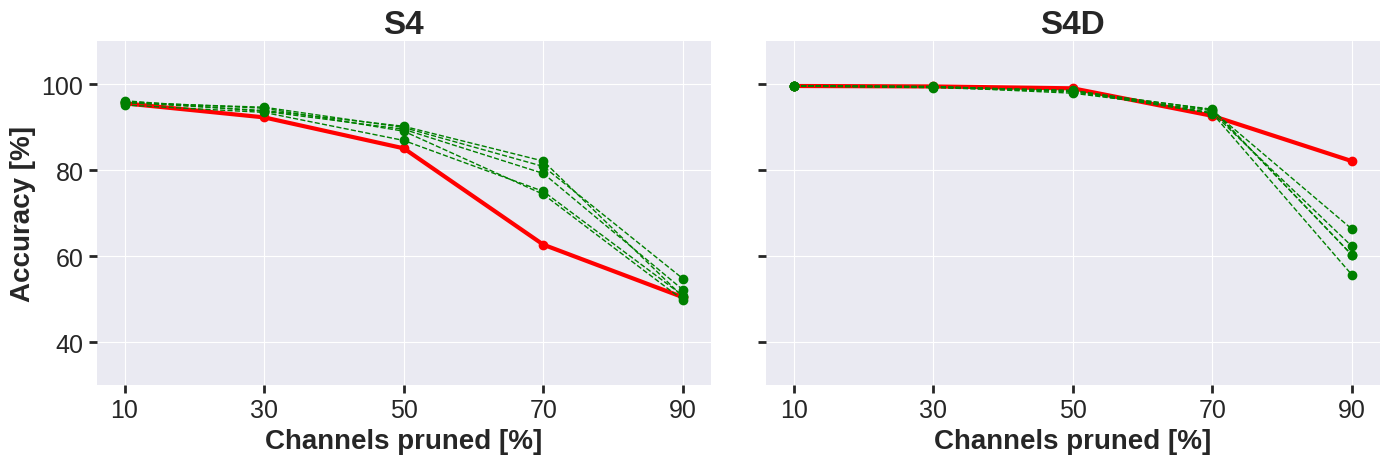}
        \caption{\textit{Pathfinder}}
        \label{fig:pathfinder_comparison}
    \end{subfigure}
    \begin{subfigure}{0.49\textwidth}
        \centering
        \includegraphics[width=0.95\linewidth]{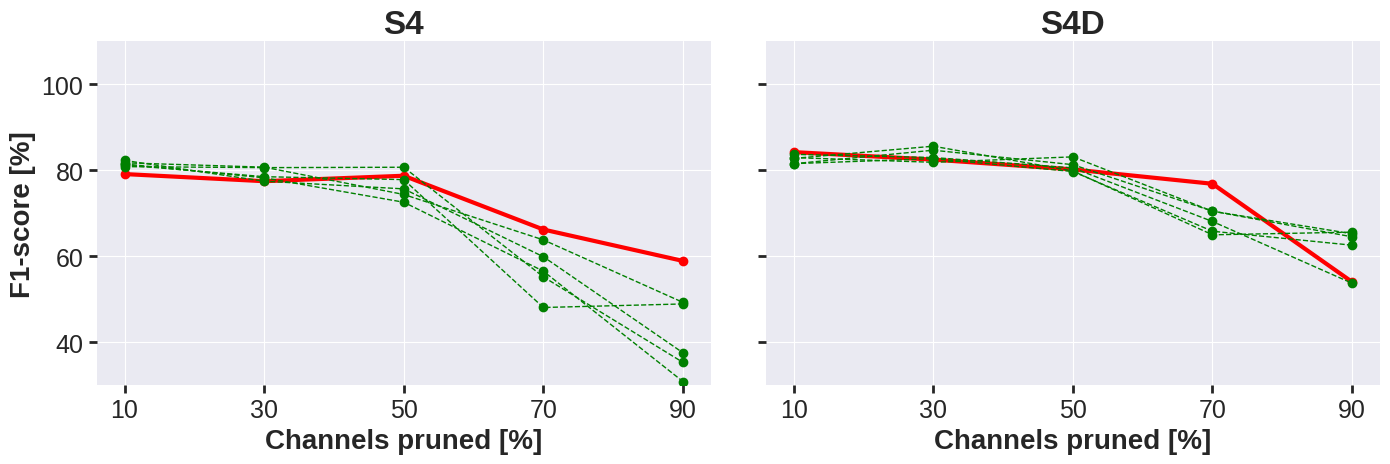}
        \caption{\textit{ECG}}
        \label{fig:ecg_comparison}
    \end{subfigure}

    \caption{\textit{Comparison between the one-shot depth-aware pruning methodology (in green) and the one-shot random channel pruning (in red).}}
    \label{fig:methodology_comparison}
    \vspace{-10pt}
\end{figure}

\subsection{Setup}

We consider three Long Range Arena (LRA) benchmarks~\cite{lra}--\textit{ListOps}, \textit{Pathfinder}, and \textit{IMDb}--covering symbolic reasoning, vision, and text, and the \textit{CODE} ECG benchmark~\cite{code}, a long-sequence biomedical classification task. These datasets provide diverse sequence modeling conditions, ranging from long-range dependency reasoning to fine-grained temporal signal analysis.

Baseline S4 and S4D architectures follow the original implementations and training settings~\cite{s4,s4d,ecg}. We evaluate five global pruning rates (10\%, 30\%, 50\%, 70\%, 90\%) and set fine-tuning to $1/8$ of the original training epochs.

Latency was measured on an NVIDIA Jetson Orin Nano 8 GB (Ampere-architecture GPU, 1024 CUDA cores, 8 GB LPDDR5 unified memory at 68 GB/s), running JetPack 6.1 in the 15 W power mode and with GPU/CPU clocks fixed at their maximum via the platform's clock-locking utility. Measurements used batch size 1 at each benchmark's full sequence length (4096 for IMDb, 5995 for ListOps, 1024 for pathfinder and 4096 for ECG), reporting the mean over 100 runs and after 50 warm-up iterations.

\begin{figure}[t]
    \begin{subfigure}{0.49\textwidth}
        \centering
        \includegraphics[width=0.95\linewidth]{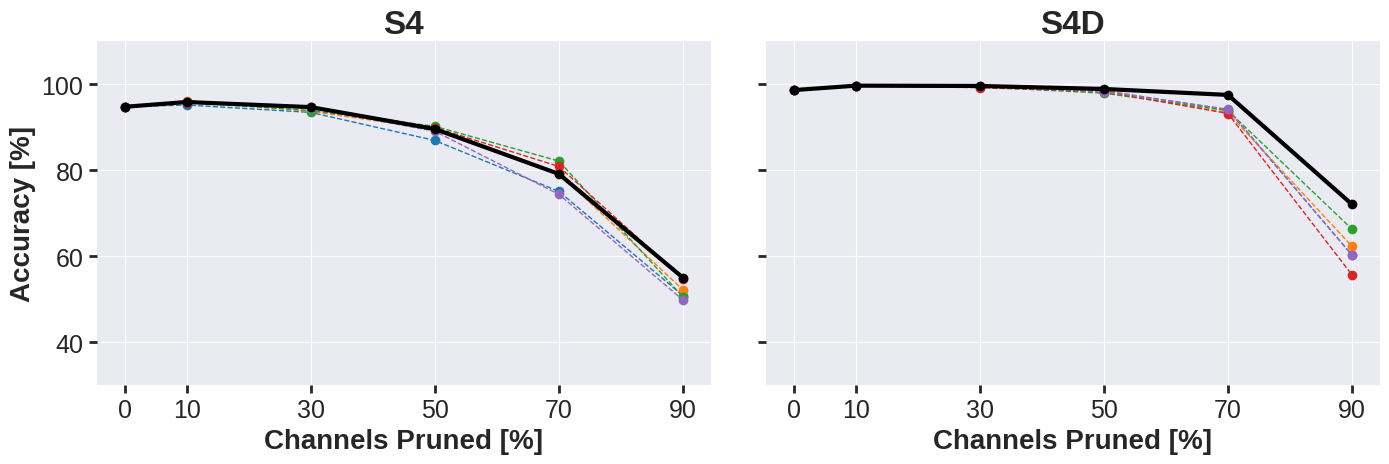}
        \caption{\textit{Pathfinder}}
        \label{fig:pathfinder_results}
    \end{subfigure}
    \begin{subfigure}{0.49\textwidth}
        \centering
        \includegraphics[width=0.95\linewidth]{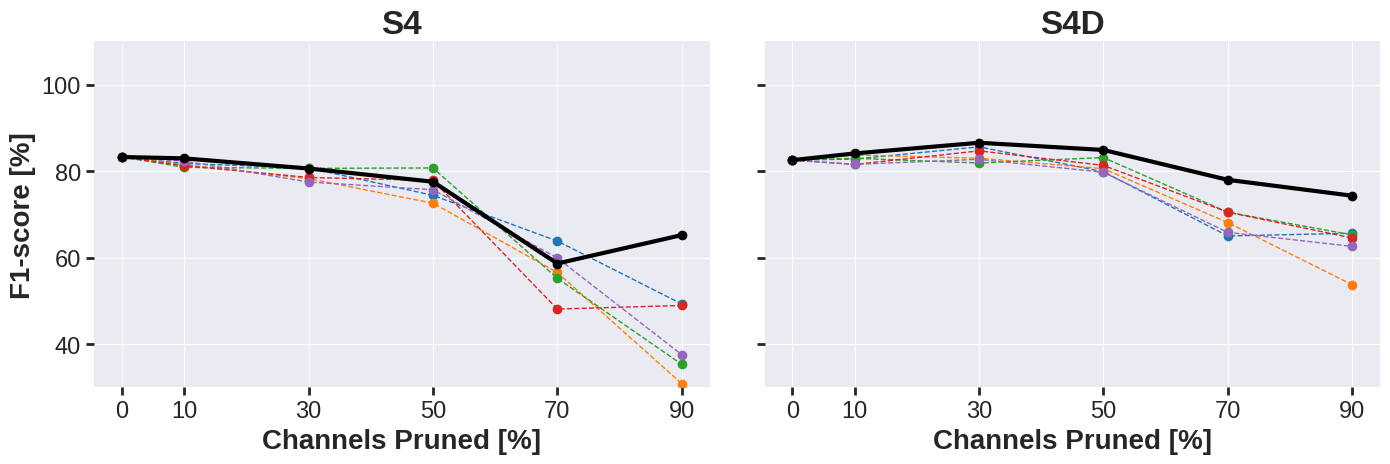}
        \caption{\textit{ECG}}
        \label{fig:ecg_results}
    \end{subfigure}

    \caption{\textit{Pruning rate vs. accuracy (F1-score for ECG). The black curve denotes our pruning strategy. Colored curves denote independent one-shot pruning experiments with different random seeds, where each step starts from the base model.}}
    \label{fig:grapichs}
    \vspace{-12pt}
\end{figure}

\subsection{Evaluated pruning strategies}

We compare four channel selection strategies: (i) layer-wise pruning, (ii) random channel pruning, (iii) uniform channel pruning, and (iv) the proposed depth-aware strategy. Table~\ref{tab:tabel_layer-wise_strategy} reports the layer-wise ablation for some model-dataset couples, where layers were selectively deactivated. Across models and datasets, pruning early layers causes the largest degradation, while deeper layers are consistently more tolerant, indicating that early layers are more critical and later ones more redundant. Furthermore, Figure~\ref{fig:methodology_comparison} highlights representative one-shot pruning results comparing depth-aware and random channel pruning, where each pruning rate is applied independently starting from the same baseline model. Depth-aware pruning consistently matches or outperforms random pruning up to medium pruning rates, while uniform pruning overlaps with random and is omitted for clarity. These results support the idea of allocating larger pruning budgets to deeper layers.

\begin{table}[ht!]
     \centering
     \caption{Results of the layer-wise pruning experiments. At each test, a single layer is entirely pruned leaving only one active channel. The column \textit{Layers Pruned} indicates the index of the layer that is removed, while \textit{italic results} denotes the unpruned baseline metric.}
     \label{tab:tabel_layer-wise_strategy}
     \scriptsize
     \begin{tabular*}{\columnwidth}{@{\extracolsep{\fill}} c c c c c c c}
        \toprule
        & \multicolumn{2}{c}{\scriptsize\textbf{ListOps}} & \multicolumn{2}{c}{\scriptsize\textbf{Pathfinder}} &
        \multicolumn{2}{c}{\scriptsize\textbf{IMDb}} \\
        \cmidrule(lr){2-3} \cmidrule(lr){4-5} \cmidrule(lr){6-7}
        \multirow{-2.5}{*}{\shortstack{\scriptsize\textbf{Layers} \\ \scriptsize\textbf{Pruned}}} & \textbf{S4} & \textbf{S4D} & \textbf{S4} & \textbf{S4D} & \textbf{S4} & \textbf{S4D} \\
        \midrule
         & \textit{61.30} & \textit{59.70} & \textit{94.77} & \textit{98.66} & \textit{84.78} & \textit{87.25}          \\
        0          & 40.25          & 59.80          & 51.13          & 98.14   & 52.07          & 66.11          \\
        1          & 49.95          & 59.10          & 83.07          & 98.80   & 84.41          & 87.18          \\
        2          & 49.40          & 59.00          & 90.46          & 99.09   & 85.20          & 87.82          \\
        3          & 58.65          & 59.65          & 94.26          & 99.41   & 84.72          & 87.74          \\
        4          & 56.75          & 59.20          & 95.33          & 99.48   & ---            & ---            \\
        5          & 58.55          & 59.20          & 96.03          & 99.53   & ---            & ---            \\
        6          & ---            & 58.60          & ---            & ---     & ---            & ---            \\
        7          & ---            & 60.05          & ---            & ---     & ---            & ---            \\
        \bottomrule
        \vspace{-15pt}
    \end{tabular*}
\end{table}

\subsection{Incremental pruning}

We compare one-shot pruning, where each pruning rate is applied independently to the baseline model, against the incremental $S4oP$ strategy, where pruning is applied progressively. As shown in Figure~\ref{fig:grapichs}, incremental pruning consistently matches or improves one-shot pruning, especially at higher pruning rates, confirming the benefit of progressive model adaptation.

\begin{table}[t!]
    \centering
    
    \caption{Metric (F1-score for ECG and accuracy for the remaining datasets), latency and model parameters results for the S4 and S4D models across all datasets.
    For the IMDb dataset, model parameters are reported excluding the embedding layer.}

    \label{tab:tabel_results_all}
    \renewcommand{\arraystretch}{0.86}
    \setlength{\tabcolsep}{8pt}
    \resizebox{\columnwidth}{!}{\begin{tabular}{l l p{1cm} p{1cm} p{2cm} p{2.8cm}}
        \textbf{Dataset} & \textbf{Model} & \textbf{Pruning Rate} & \textbf{Metric} & \textbf{Latency[ms] (speedup\%)} & \textbf{\#parameters (reduction\%)} \\
        \toprule
        \multirow{12}{*}{IMDb} 
        & \multirow{6}{*}{S4} 
        & 0\% & \textit{84.78} & \textit{145.4} & \textit{99,970} \\
        & & 10\% & 84.59 & 132.0\hfill(-9.2\%)  & 93,262\hfill(-6.7\%) \\
        & & 30\% & 84.35 & 104.6\hfill(-28.1\%) & 80,104\hfill(-19.9\%) \\
        & & 50\% & 83.66 & 78.5\hfill(-46.0\%) & 66,946\hfill(-33.0\%) \\
        & & 70\% & 81.48 & 51.5\hfill(-64.6\%) & 53,788 \hfill(-46.2\%) \\
        & & 90\% & 76.93 & 25.5\hfill(-82.5\%)  & 40,630 \hfill(-59.4\%) \\
        \cmidrule(l){2-6} 
        & \multirow{6}{*}{S4D} 
        & 0\% & \textit{87.25} & \textit{146.2} & \textit{992,770} \\
        & & 10\% & \textbf{87.64} & 132.8\hfill(-9.1\%)  & 972,750\hfill(-2.0\%) \\
        & & 30\% & \textbf{87.73} & 105.2\hfill(-28.0\%) & 932,970\hfill(-6.0\%) \\
        & & 50\% & \textbf{87.66} & 78.3\hfill(-46.4\%) & 892,930\hfill(-10.0\%) \\
        & & 70\% & \textbf{87.90} & 51.1\hfill(-65.0\%) & 853,020\hfill(-14.1\%) \\
        & & 90\% & 86.58 & 23.3\hfill(-84.1\%) & 813,110\hfill(-18.1\%) \\
        \midrule 
        \multirow{12}{*}{ListOps} 
        & \multirow{6}{*}{S4} 
        & 0\% & \textit{61.30} & \textit{621.7} & \textit{401,418} \\
        & & 10\% & 59.80 & 561.8\hfill(-9.6\%)  & 381,552\hfill(-5.0\%) \\
        & & 30\% & 57.05 & 442.0\hfill(-28.9\%) & 342,078\hfill(-14.8\%) \\
        & & 50\% & 54.75 & 321.0\hfill(-48.4\%) & 302,346\hfill(-24.7\%) \\
        & & 70\% & 47.20 & 201.8\hfill(-67.5\%) & 262,614\hfill(-34.6\%) \\
        & & 90\% & 40.05 & 83.4\hfill(-86.6\%) & 223,140\hfill(-44.4\%) \\
        \cmidrule(l){2-6} 
        & \multirow{6}{*}{S4D} 
        & 0\% & \textit{59.70} & \textit{170.4} & \textit{402,954} \\
        & & 10\% & \textbf{60.25} & 154.6\hfill(-9.3\%) & 389,564\hfill(-3.3\%) \\
        & & 30\% & 58.75 & 123.1\hfill(-27.8\%) & 363,044\hfill(-9.9\%) \\
        & & 50\% & 55.30 & 91.4\hfill(-46.4\%) & 336,394\hfill(-16.5\%) \\
        & & 70\% & 49.35 & 59.6\hfill(-65.1\%) & 309,874\hfill(-23.1\%) \\
        & & 90\% & 43.30 & 26.4\hfill(-84.5\%) & 283,224\hfill(-29.7\%) \\
        \midrule 
        \multirow{12}{*}{Pathfinder} 
        & \multirow{6}{*}{S4} 
        & 0\% & \textit{94.77} & \textit{231.0} & \textit{1,190,402} \\
        & & 10\% & \textbf{95.89} & 209.3\hfill(-9.4\%)  & 1,150,670\hfill(-3.3\%) \\
        & & 30\% & 94.70 & 166.0\hfill(-28.1\%) & 1,071,722\hfill(-10.0\%) \\
        & & 50\% & 89.60 & 125.0\hfill(-45.9\%) & 992,258\hfill(-16.7\%) \\
        & & 70\% & 79.16 & 82.1\hfill(-64.5\%) & 913,052\hfill(-23.3\%) \\
        & & 90\% & 55.02 & 41.3\hfill(-82.1\%) & 833,846\hfill(-30.0\%) \\
        \cmidrule(l){2-6} 
        & \multirow{6}{*}{S4D} 
        & 0\% & \textit{98.66} & \textit{35.3} & \textit{993,794} \\
        & & 10\% & \textbf{99.69} & 32.1\hfill(-9.3\%)  & 973,774\hfill(-2.0\%) \\
        & & 30\% & \textbf{99.62} & 25.3\hfill(-28.4\%) & 933,994\hfill(-6.0\%) \\
        & & 50\% & \textbf{98.92} & 18.5\hfill(-47.7\%) & 893,954\hfill(-10.1\%) \\
        & & 70\% & 97.54 & 12.7\hfill(-64.0\%) & 854,044\hfill(-14.1\%) \\
        & & 90\% & 72.07 & 11.9\hfill(-66.3\%) & 814,134\hfill(-18.1\%) \\
        \midrule 
        \multirow{12}{*}{ECG} 
        & \multirow{6}{*}{S4} 
        & 0\% & \textit{83.25} & \textit{283.0} & \textit{267,654} \\
        & & 10\% & 82.93 & 256.3\hfill(-9.5\%)  & 254,496\hfill(-4.9\%) \\
        & & 30\% & 80.59 & 201.1\hfill(-28.9\%) & 228,180\hfill(-14.8\%) \\
        & & 50\% & 77.53 & 147.8\hfill(-47.8\%) & 201,606\hfill(-24.7\%) \\
        & & 70\% & 58.62 & 94.1\hfill(-66.8\%) & 175,290\hfill(-34.5\%) \\
        & & 90\% & 65.19 & 40.7\hfill(-85.6\%) & 148,974\hfill(-44.3\%) \\
        \cmidrule(l){2-6} 
        & \multirow{6}{*}{S4D} 
        & 0\% & \textit{82.53} & \textit{52.7} & \textit{202,118} \\
        & & 10\% & \textbf{84.07} & 47.3\hfill(-10.3\%)  & 195,488\hfill(-3.3\%) \\
        & & 30\% & \textbf{86.54} & 37.3\hfill(-29.3\%) & 182,228\hfill(-9.8\%) \\
        & & 50\% & \textbf{84.88} & 27.7\hfill(-47.4\%) & 168,838\hfill(-16.5\%) \\
        & & 70\% & 77.97 & 18.1\hfill(-65.6\%) & 155,578\hfill(-23.0\%) \\
        & & 90\% & 74.30 & 9.0\hfill(-82.9\%) & 142,318\hfill(-29.6\%) \\
        \bottomrule
        \vspace{-20pt}
    \end{tabular}}
\end{table}

\subsection{Accuracy-efficiency trade-off}

Figure~\ref{fig:grapichs} and Table~\ref{tab:tabel_results_all} summarize the main results. Across tasks, pruning up to approximately 30\% of channels generally preserves baseline accuracy, whereas more aggressive pruning leads to task-dependent performance degradation.

S4D is consistently more robust than S4, often matching or exceeding baseline accuracy under moderate pruning. This trend is particularly evident on \textit{Pathfinder} and \textit{ECG}, where S4D remains stable under aggressive pruning and in some cases improves over the baseline. On \textit{IMDb}, both architectures are highly robust, maintaining near-baseline accuracy up to 70\% pruning, indicating substantial redundancy. In contrast, \textit{ListOps} is the most pruning-sensitive benchmark.

\begin{figure}[ht!]
    \centering
    \includegraphics[width=0.95\linewidth]{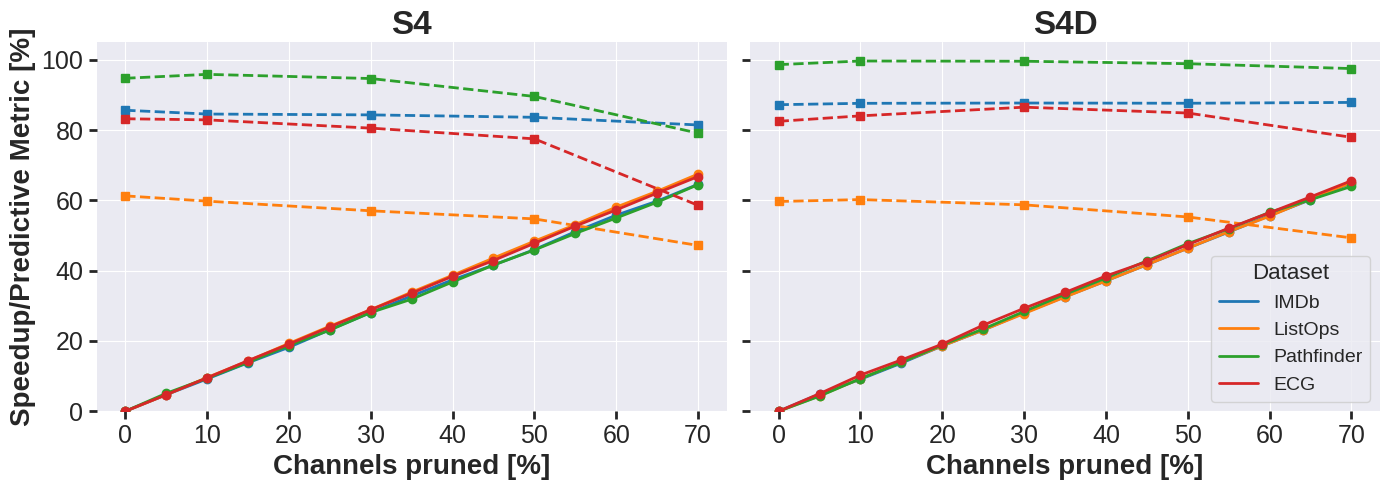}
    \caption{\textit{Speedup and accuracy across pruning rates on a Jetson Orin Nano device. Continuous lines show inference latency reduction, while dashed lines indicate predictive accuracies for each dataset-model configuration}}
    \label{fig:speedup_jetson}
    \vspace{-8pt}
\end{figure}

Figure~\ref{fig:speedup_jetson} shows that pruning consistently reduces inference latency, with speedups that scale nearly proportionally to the fraction of removed channels. Across datasets, moderate pruning already yields substantial gains, while aggressive pruning often reduces latency by 40-60\%, depending on model and task. Parameter count follows the same monotonic trend, although reductions are smaller than latency gains since pruning affects only SSM operators.

These results show that the operator-level pruning we presented provides a favorable accuracy-efficiency trade-off: moderate pruning often preserves, and in some cases improves, predictive performance while reducing inference cost.


\section{Conclusion}\label{SEC:CONCLUSION}
We presented a systematic study of pruning S4- and S4D-based models through a unified framework for training, pruning, and evaluation. Experiments across multiple datasets show that pruning up to 30\% of model operators preserves accuracy while reducing inference latency and parameter count. We further observe a depth-dependent redundancy pattern: early layers are more sensitive to pruning, whereas deeper layers tolerate more aggressive operator removal. This trend is consistent across datasets and architectures and guides effective pruning strategies for sequence models.


\newpage
\bibliographystyle{IEEEtran}
\bibliography{citations.bib}

\end{document}